\documentclass[letterpaper]{article} 
\usepackage{aaai2026}  
\usepackage{times}  
\usepackage{helvet}  
\usepackage{courier}  
\usepackage[hyphens]{url}  
\usepackage{graphicx} 
\usepackage{natbib}  
\usepackage{caption} 
\usepackage{subcaption}
\usepackage{multirow}
\usepackage{mathrsfs}
\usepackage{booktabs}
\usepackage{amsmath,amsfonts,amssymb}
\usepackage{float}
\usepackage{xcolor}
\usepackage{algorithm}
\usepackage{algorithmic}

\usepackage{newfloat}
\usepackage{listings}
\DeclareCaptionStyle{ruled}{labelfont=normalfont,labelsep=colon,strut=off} 
\floatstyle{ruled}
\newfloat{listing}{tb}{lst}{}
\floatname{listing}{Listing}
\title{Perceptual Quality Assessment of 3D Gaussian Splatting: A Subjective Dataset and Prediction Metric}

\author{
    Zhaolin Wan\textsuperscript{\rm 1},
    Yining Diao\textsuperscript{\rm 1},
    Jingqi Xu\textsuperscript{\rm 1},
    Hao Wang\textsuperscript{\rm 1,3},\\ 
    Zhiyang Li\textsuperscript{\rm 2},
    Xiaopeng Fan\textsuperscript{\rm 1,4}\thanks{Corresponding Author.}, 
    Wangmeng Zuo\textsuperscript{\rm 1},
    Debin Zhao\textsuperscript{\rm 1}
}
\affiliations{
    \textsuperscript{\rm 1}Faculty of Computing, Harbin Institute of Technology, Harbin, China\\
    \textsuperscript{\rm 2}School of Information Science and Technology, Dalian Maritime University, Dalian, China\\
    \textsuperscript{\rm 3}Department of Computer Science, City University of Hong Kong, Hong Kong\\
    \textsuperscript{\rm 4}Suzhou Research Institute, Harbin Institute of Technology, Suzhou, China\\
    
    \{zlwan, fxp, dbzhao\}@hit.edu.cn, 
    \{23S003005, 24S103369\}@stu.hit.edu.cn,\\
    ho.wang@my.cityu.edu.hk, 
    lizy0205@dlmu.edu.cn, 
    cswmzuo@gmail.com
}

\usepackage{bibentry}

\begin{document}

\maketitle

\begin{abstract}
With the rapid advancement of 3D visualization, 3D Gaussian Splatting (3DGS) has emerged as a leading technique for real-time, high-fidelity rendering. While prior research has emphasized algorithmic performance and visual fidelity, the perceptual quality of 3DGS-rendered content, especially under varying reconstruction conditions, remains largely underexplored. In practice, factors such as viewpoint sparsity, limited training iterations, point downsampling, noise, and color distortions can significantly degrade visual quality, yet their perceptual impact has not been systematically studied. To bridge this gap, we present 3DGS-QA, the first subjective quality assessment dataset for 3DGS. It comprises 225 degraded reconstructions across 15 object types, enabling a controlled investigation of common distortion factors. Based on this dataset, we introduce a no-reference quality prediction model that directly operates on native 3D Gaussian primitives, without requiring rendered images or ground-truth references. Our model extracts spatial and photometric cues from the Gaussian representation to estimate perceived quality in a structure-aware manner. We further benchmark existing quality assessment methods, spanning both traditional and learning-based approaches. Experimental results show that our method consistently achieves superior performance, highlighting its robustness and effectiveness for 3DGS content evaluation. The dataset and code are made publicly available at \url{https://github.com/diaoyn/3DGSQA} to facilitate future research in 3DGS quality assessment.

\end{abstract}

\section{Introduction}
In recent years, three-dimensional (3D) visual representations have gained significant attention across a wide range of applications, including virtual reality, gaming, medical imaging, and digital twins \cite{fei20243d}. These technologies enable immersive and interactive experiences. Among emerging techniques, 3D Gaussian Splatting (3DGS) \cite{kerbl20233d} has recently stood out as a compelling alternative to Neural Radiance Fields (NeRF) \cite{mildenhall2021nerf}. By leveraging anisotropic 3D Gaussians that can be directly rasterized via a tile-based renderer, 3DGS bypasses the computational overhead of ray marching used in NeRF. This innovation delivers real-time rendering with competitive visual fidelity, while significantly reducing both training and inference time.

Despite its growing adoption and efficiency, existing research on 3DGS has largely focused on objective performance metrics \cite{bao20253d}, such as rendering speed \cite{lee2024gscore, durvasula2023distwar}, memory efficiency \cite{girish2024eagles, navaneet2023compact3d}, and geometric accuracy \cite{jiang2024gaussianshader, liu2024mirrorgaussian}. In contrast, perceptual quality—how humans actually perceive 3DGS-rendered content—remains largely underexplored. Reconstruction degradations from sparse input views, limited optimization, or photometric noise can introduce artifacts like blur, transparency inconsistency, and geometric distortion, which significantly degrade user experience. Although perceptual quality assessment has been extensively studied for 2D images \cite{wang2004image, wang2023exploring}, point clouds \cite{alexiou2018point, liu2023point}, and meshes \cite{abouelaziz2020no, zhang2022no}, these methods generally rely on explicit geometric models or 2D projections. These assumptions do not transfer well to 3DGS, which adopts a probabilistic volumetric representation and a rasterization-based rendering process. Thus, existing methods are ill-suited to assess the perceptual integrity of 3DGS-rendered content.

Furthermore, while 3DGS-specific quality assessment datasets are crucial, existing efforts remain limited in both scope and generality. For example, Zhang et al.\cite{zhang2025evaluating} and Martin et al.\cite{martin2025gs} conducted comparative studies on a few 3DGS models but lack fine-grained control over distortion types and severity levels, hindering systematic analysis of perceptual degradation mechanisms. Yang et al. \cite{yang2024benchmark} targets only compression-induced artifacts using a graph-based framework, overlooking other critical factors such as input sparsity, under-training, or additive noise. Consequently, existing datasets only partially cover the range of relevant distortions and fall short of enabling controlled, comprehensive perceptual studies. This lack of comprehensive benchmarks hinders the development, evaluation, and deployment of perceptually-driven quality assessment models and rendering strategies for 3DGS.

To address this gap, we present the first controlled and systematic study on perceptual quality assessment for 3D Gaussian Splatting. Specifically, we introduce 3DGS-QA, the first subjective quality assessment dataset for 3DGS. It consists of 225 distorted models across 15 object categories, each reconstructed under controlled variations in key rendering parameters, enabling in-depth analysis of how these factors influence human perception. Furthermore, we propose GSOQA, a no-reference quality prediction model that directly operates on the native 3D Gaussian primitives—bypassing the need for 2D projections or reference data. By extracting spatial and photometric features from the Gaussians, the model accurately predicts perceived quality using a deep neural architecture. Lastly, we establish a comprehensive evaluation protocol based on 3DGS-QA and benchmark our approach against state-of-the-art methods. Experimental results show that our method consistently outperforms existing baselines. In summary, our main contributions are as follows:

\begin{itemize}
\item We construct 3DGS-QA, the first subjective quality assessment dataset for 3D Gaussian Splatting, featuring 225 models across 15 object types, with controlled distortions to study perceptual degradation factors.
\item We propose GSOQA, a no-reference quality prediction model that directly operates on native 3D Gaussian primitives to extract spatial and photometric cues, enabling prediction without 2D projections or references.
\item We develop a benchmark evaluation protocol based on 3DGS-QA and show that our method significantly outperforms both traditional and learning-based quality metrics in assessing 3DGS-generated content.
\item We provide an in-depth comparative study of existing quality metrics applied to 3DGS content, highlighting their limitations and offering insights into the challenges of perceptual modeling for 3DGS rendering paradigms.
\end{itemize}

\section{3DGS Database Construction and Analysis}\label{sec:3dgsdatabase}
\subsection{3DGS Model Construction}
To facilitate perceptual quality analysis of 3DGS models, we designed a rigorous and reproducible pipeline comprising object selection, multi-view rendering, point cloud generation, and 3DGS synthesis. 

\subsubsection{Object Selection and Categorization} 
\begin{figure}[ht]
\centering
\includegraphics[width=0.45\textwidth]{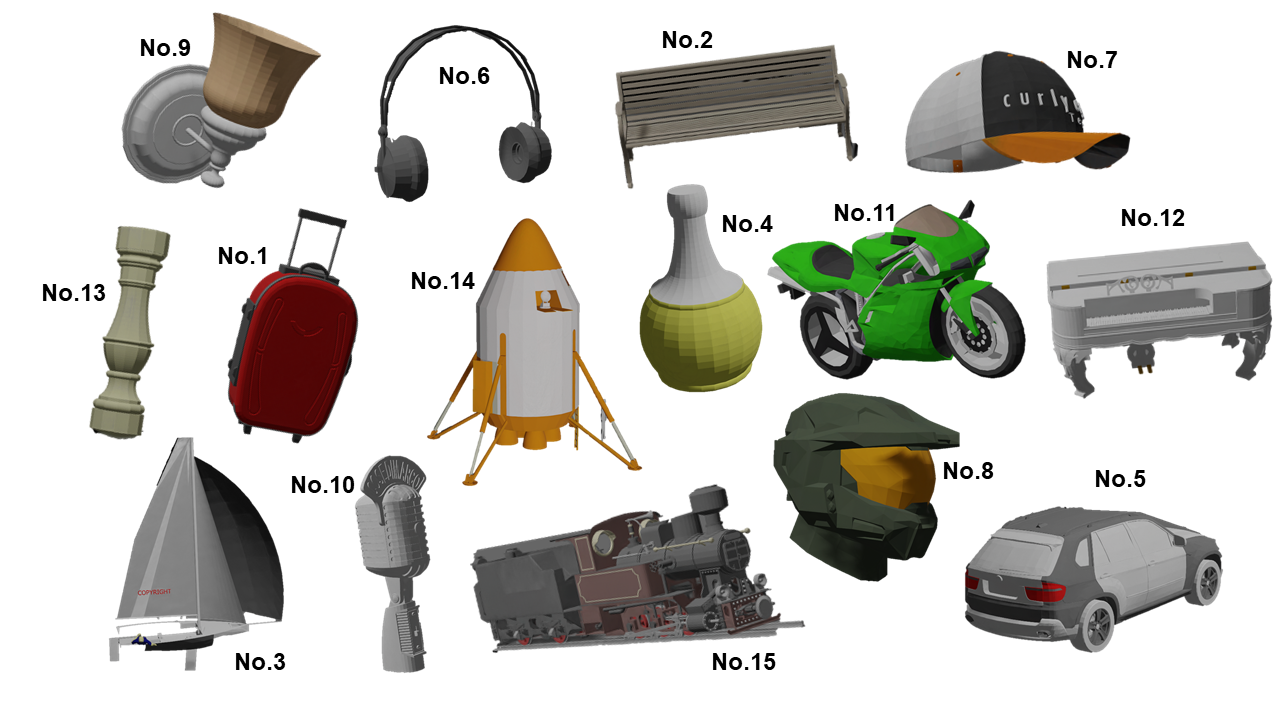}%
\caption{The source models of our database 3DGS-QA.}
\label{fig:3DGS}
\end{figure}

We selected 15 representative object categories from two widely-used 3D datasets: ShapeNet \cite{chang2015shapenet} and ModelNet \cite{wu20153d}, to ensure diversity in geometry, topology, and semantic function. The chosen categories span four semantic groups: transportation vehicles, personal accessories, audio devices and instruments, and structural or utility items.
As visualized in Fig.~\ref{fig:3DGS}, although the models are synthetic, this setup allows for fine-grained control over distortion parameters, consistent ground-truth availability, and experimental reproducibility, which are otherwise challenging to ensure in real-world scans due to noise and scene variability.

\subsubsection{Multi-view Rendering via Blender}
To generate high-quality multi-view image data, we utilized Blender for physically-based rendering, following the NeRF setup \cite{mildenhall2021nerf}. We configured a parallel directional light to simulate natural illumination, and enabled realistic shading to preserve surface textures. Cameras were uniformly placed on a spherical shell around each object, with parameters (focal length, aperture, shutter speed) adjusted based on object scale to ensure clear and consistent perspectives. All images were rendered at a resolution of 1200×1200 pixels with a pure white background to support accurate segmentation and subsequent 3DGS reconstruction.

\begin{figure*}[!t]
    \centering
    \begin{minipage}{0.18\textwidth}
        \centering
        \includegraphics[width=\textwidth]{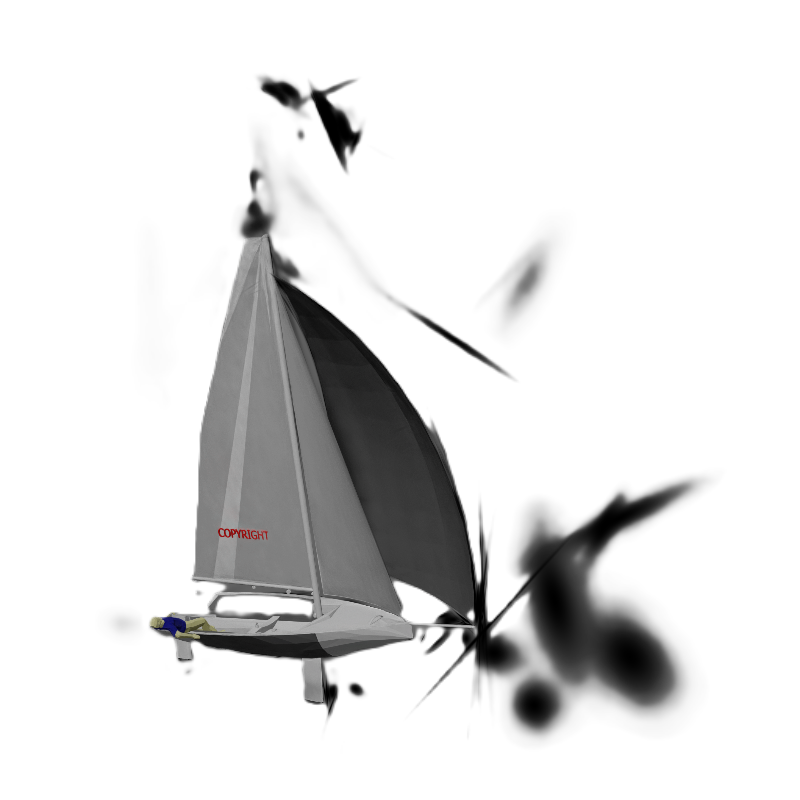} \\
        (a)
    \end{minipage}
    \begin{minipage}{0.18\textwidth}
        \centering
        \includegraphics[width=\textwidth]{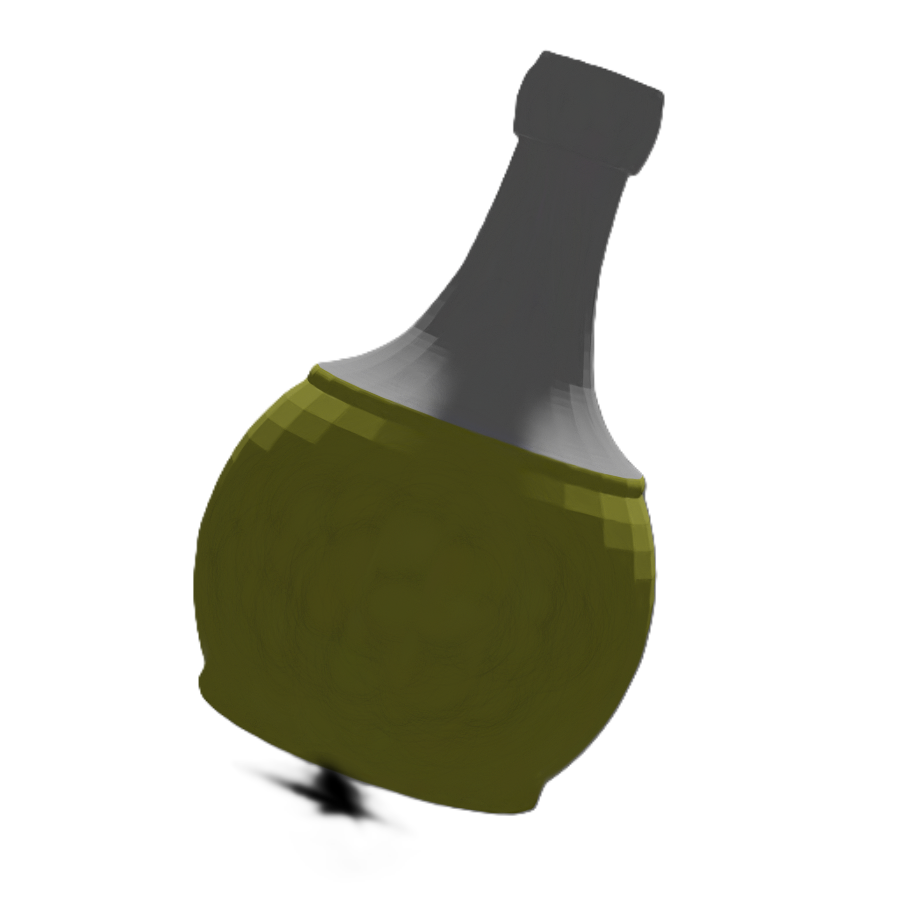} \\
        (b)
    \end{minipage}
    \begin{minipage}{0.18\textwidth}
        \centering
        \includegraphics[width=\textwidth]{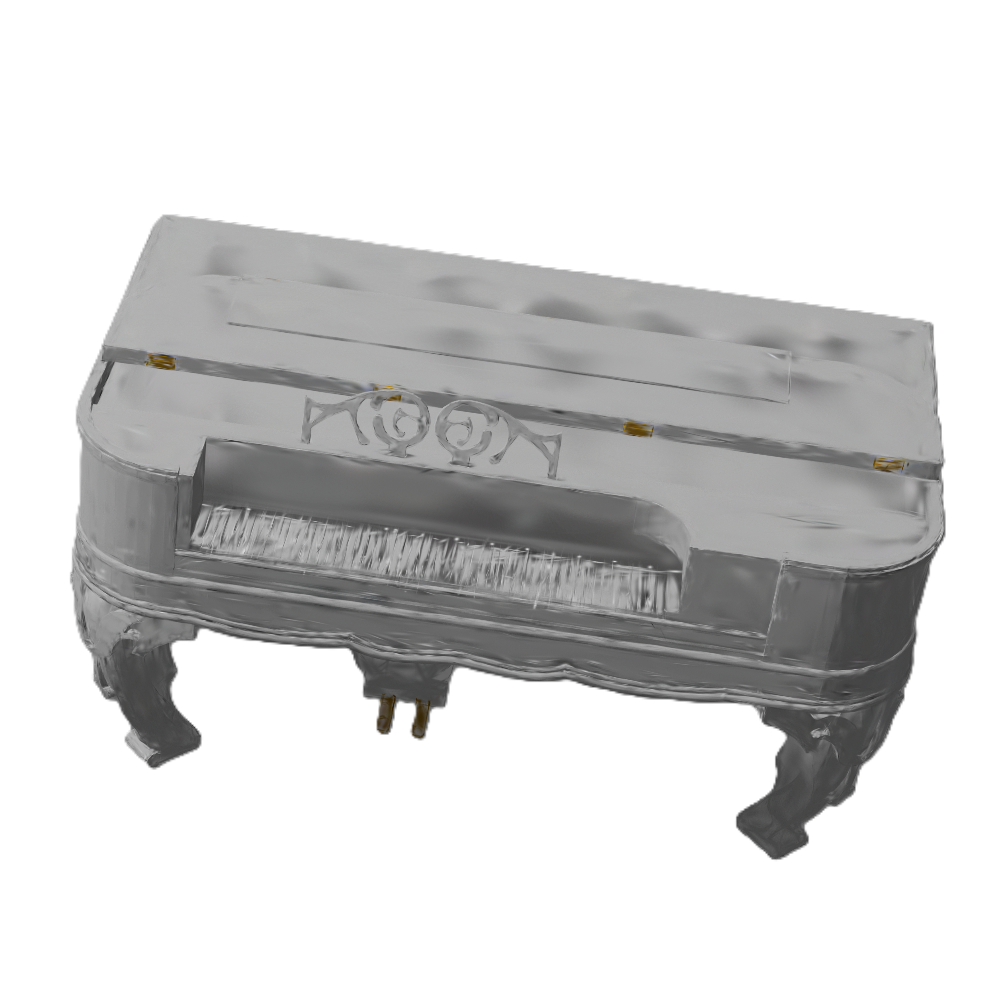} \\
        (c)
    \end{minipage}
    \begin{minipage}{0.18\textwidth}
        \centering
        \includegraphics[width=\textwidth]{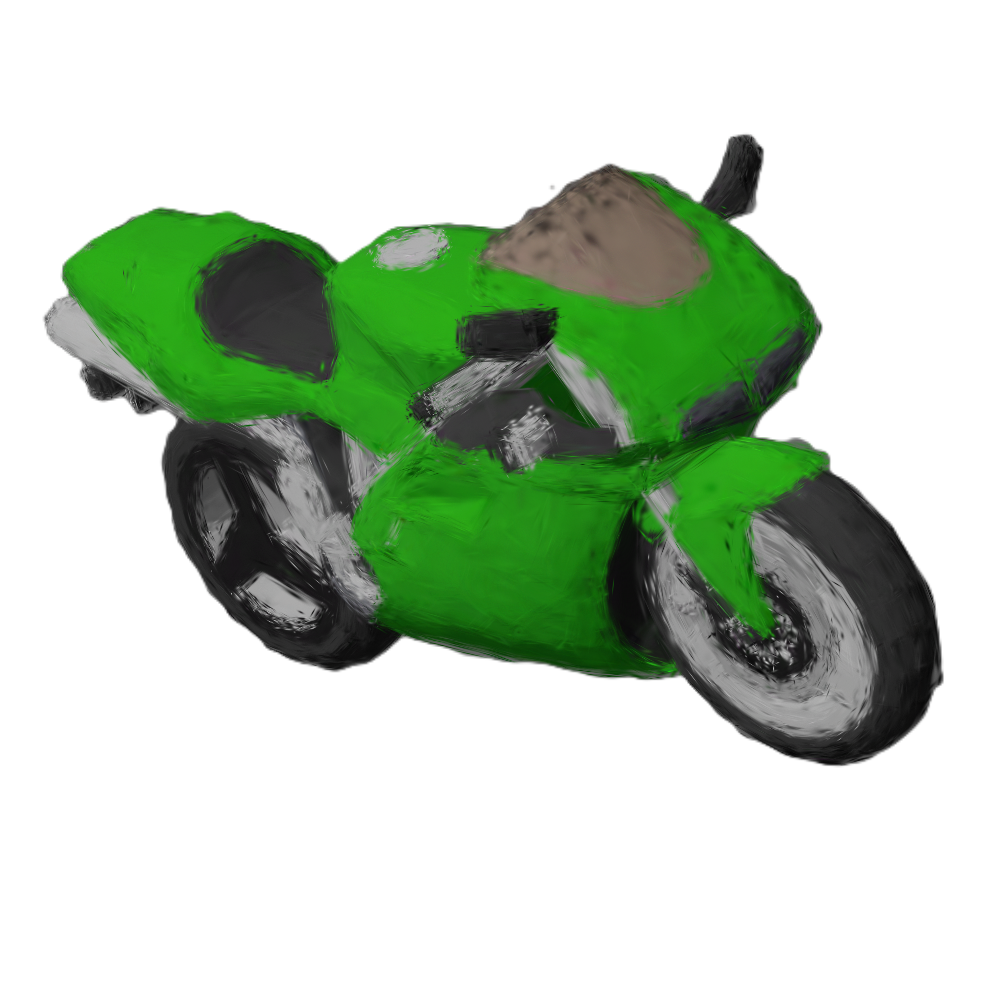} \\
        (d)
    \end{minipage}
    \begin{minipage}{0.18\textwidth}
        \centering
        \includegraphics[width=\textwidth]{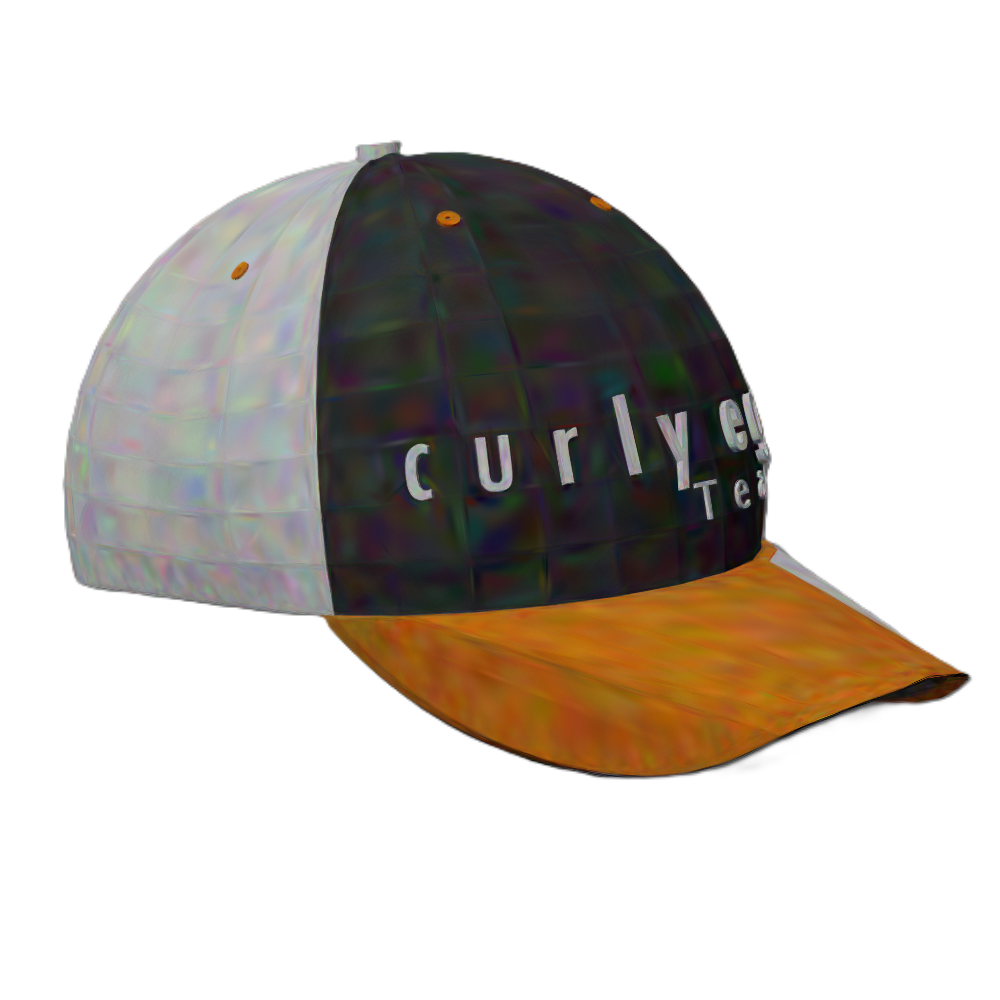} \\
        (e)
    \end{minipage}
    \caption{Examples of distortion types in 3DGS models: (a) reduced viewpoints, (b) limited training, (c) point downsampling, (d) spatial noise, and (e) color perturbation.}
    \label{fig:distort}
\end{figure*}

\subsubsection{Point Cloud Initialization using COLMAP}
With rendered images and calibrated camera poses, we utilized COLMAP \cite{schonberger2016structure} to generate initial point clouds. Specifically, we adopted mesh-guided uniform sampling to extract 5,000 surface points per model, allocating points to each triangular face proportionally to its area. This strategy ensures even spatial distribution and preserves geometric detail. The approach aligns with standard practices in ShapeNet and ModelNet, enhancing compatibility and reproducibility.

\subsubsection{3D Gaussian Splatting Synthesis}
Based on the generated point clouds and multi-view imagery, we constructed 3DGS models using the official implementation by Kerbl et al. \cite{kerbl20233d}, employing the following stages:

\begin{itemize}
\item Preprocessing: Initial image denoising and point cloud refinement were applied to enhance alignment accuracy and structural completeness.
\item Gaussian Parameter Optimization: Each point was modeled as a Gaussian primitive with learnable attributes, including position, anisotropic scale, opacity, orientation, and color. These parameters were jointly optimized to minimize multi-view photometric reconstruction error.

\item Final Model Generation: The optimized set of Gaussians produces a dense, view-consistent representation that faithfully reconstructs both geometric structure and visual appearance of the original object.
\end{itemize} 

The characteristics of the resulting 3DGS models are summarized in Table~\ref{tab:ablation}.

\begin{table}[ht]
    \centering
    \caption{Summary of 3DGS models included in our dataset.}
    \resizebox{0.45\textwidth}{!}{%
    \begin{tabular}{cccc}
\toprule
Index & Name & GS points & Description \\
\midrule
1 & Bag & 167, 723 & Personal accessories \\
2 & Bench & 422, 245 & Utility items \\
3 & Boat & 137, 654 & Transportation vehicles \\
4 & Bottle & 154, 380 & Personal accessories \\
5 & Car & 270, 952 & Transportation vehicles \\
6 & Earphone & 154, 106 & Audio devices  \\
7 & Hat & 105, 988 & Personal accessories \\
8 & Helmet & 131, 292 & Personal accessories \\
9 & Lamp & 178, 853 & Utility items \\
10 & Microphone & 242, 014 & Audio devices \\
11 & Motorbike & 206, 719 & Transportation vehicles \\
12 & Piano & 335, 169 & Instruments \\
13 & Pillar & 188, 964 & Structural items\\
14 & Rocket & 182, 997 & Transportation vehicles \\
15 & Train & 197, 414 & Transportation vehicles \\
\bottomrule
\end{tabular}
    }
    \label{tab:ablation}
\end{table}

\subsection{Distortion Generation}

Besides our implementation, different 3DGS pipelines may exhibit distinct degradation patterns, termed model-related distortions, as explored by Martin et al. (2025). In contrast, this work focuses on perceptual distortions such as reduced view coverage, limited training, and noise, which are common across various 3DGS pipelines and not tied to specific implementations. To systematically simulate real-world challenges in data acquisition, processing, and rendering, we introduce five representative distortion types. 

\subsubsection{Reduced Viewports (Reconstruction Distortion)}
To investigate the impact of limited viewpoints, we uniformly reduce the number of viewpoints (360, 270, 180) to simulate regular angular sampling (1°, 1.5°, 2°) in real-world scans. A reduced number of views results in sparser image coverage and weaker geometry-texture correspondence, effectively simulating practical constraints such as limited sensor mobility or reduced acquisition time.

\subsubsection{Limited Training (Reconstruction Distortion)}
To simulate under-converged model optimization, we train 3DGS models with two levels of iteration counts: 7,000 and 30,000. Insufficient training iterations can lead to suboptimal fitting of Gaussian primitives to image features, often producing blurred or structurally incomplete models.

\subsubsection{Point Downsampling (Synthesis Distortion)}
To emulate resolution degradation, Gaussian primitives are randomly downsampled to 25\%, 50\%, and 75\% of their original count. A Bernoulli sampling strategy with Poisson-disk constraints is applied to avoid excessive clustering. The minimum inter-point distance is computed as: $r_{\text{min}} = \sqrt[3]{\frac{V}{N\times p}}$, where $V$ is the scene’s bounding box volume, $N$ is the total number of Gaussian primitives, and $p$ is the sampling probability.

\subsubsection{Spatial Gaussian Noise (Synthesis Distortion)}
To simulate sensor-induced positional noise, Gaussian perturbations are added to each primitive’s centroid. Specifically, for each point $P_i \in R^3$, the perturbed position is defined as:
\begin{equation}
  \mathbf{P}'_i = \mathbf{P}_i + \boldsymbol{\epsilon}, \quad \boldsymbol{\epsilon} \sim \mathcal{N}(0, \sigma^2\mathbf{I}_3),
\end{equation}
with $\sigma \in \{0.001, 0.005, 0.01\}$ controlling the noise severity.

\subsubsection{Color Harmonic Perturbation (Synthesis Distortion)}
To model appearance degradation, uniform noise is added to each spherical harmonic coefficient $SH_i$ of the Gaussians. The perturbed value is computed as:
\begin{equation}
    SH_i' = SH_i + \Delta SH_i, \quad \Delta SH_i \sim \mathcal{U}(-\delta, \delta),
\end{equation}
with $\delta \in \{0.01, 0.05, 0.1\}$ determining the variation level.


Overall, each of the 15 base 3DGS models is subjected to 2 reconstruction and 3 synthesis distortion types (each with 3 severity levels), generating 15×(2×3+3×3)=225 distinctly degraded models. Fig.~\ref{fig:distort} shows representative examples. The proposed 3DGS-QA dataset thus provides a comprehensive and fine-grained basis for perceptual quality assessment across distortion types and severity levels.

\subsection{Subjective Experiment}
To assess the quality of distorted 3DGS models, we conducted a controlled subjective experiment using a standardized virtual viewing environment and scoring protocol.

\subsubsection{Rendering and Viewing Setup}
Distorted 3DGS models were rendered and displayed in Unity3D within an interactive scene. A virtual camera followed a circular orbit, capturing 360 uniformly spaced viewpoints around the object. The camera radius was automatically adjusted to ensure full visibility of each model, with the focus aligned to its geometric center. Each rendering sequence was exported at 960×960 resolution, compiled into 6-second videos at 60 fps, providing smooth, occlusion-free coverage of the full 360° view.

\begin{figure}[!t]
\centering
\includegraphics[width=0.45\textwidth]{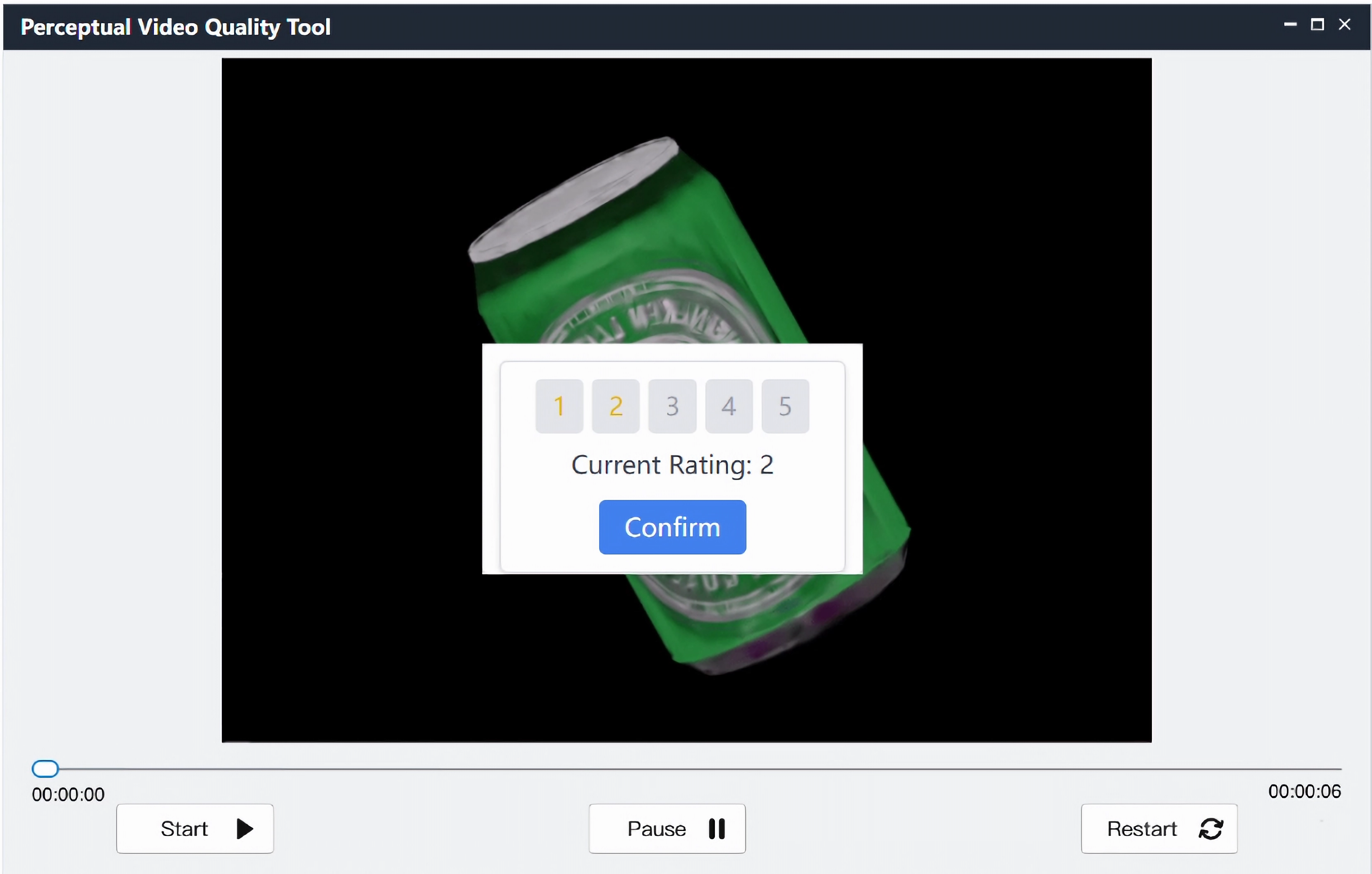}%
\caption{The subjective quality assessment interface.}
\label{fig:PVQT}
\end{figure}

\subsubsection{Experimental Protocol}
We adopted the Absolute Category Rating with Hidden Reference (ACR-HR) protocol, following the ITU-R BT.500-15 standard. The hidden reference for each distortion type was a high-quality baseline model reconstructed from 360 clean views and trained for 30,000 iterations. Participants viewed a randomized sequence of video stimuli on a calibrated 24-inch monitor (1920×1080), under standardized lighting in a quiet indoor environment.

As shown in Fig.~\ref{fig:PVQT}, participants could play, pause, and replay videos through an interactive interface. All objects were rendered against a black background to enhance contrast and highlight subtle artifacts. After viewing each video, participants rated the perceived quality on a 5-point scale (1 = Bad, 5 = Excellent).

\subsubsection{Participants and Training}
Thirty-three participants (23 male, 10 female; ages 19–35), all with normal or corrected vision and no known color vision deficiencies, took part in the study. Before the formal assessment, a training session familiarized them with the interface and rating scale using sample models with varied distortions, helping reduce learning bias and rating variance. Each participant rated all 225 video sequences (6 seconds each). The full session, including training and breaks, lasted 40–60 minutes.

To ensure data reliability, we applied an outlier rejection procedure. Ratings deviating significantly from the interquartile range of corresponding Mean Opinion Scores (MOS) were flagged. Participants with more than 5\% outlier ratings or consistent scoring bias (e.g., extreme or uniform ratings) were excluded. Final MOS values were computed by averaging valid participant scores for each model.

\begin{figure}[!t]
\centering
\includegraphics[width=0.45\textwidth]{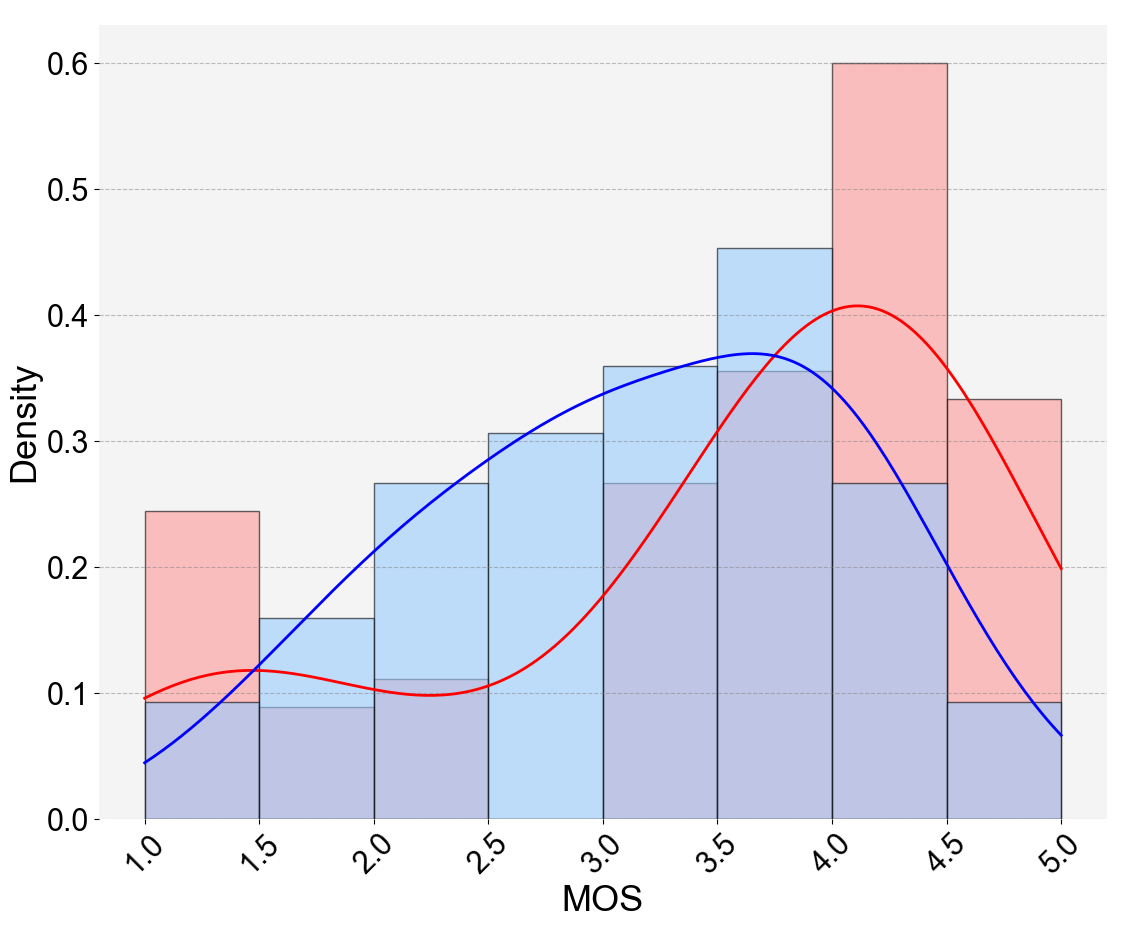}%
\caption{MOS distribution across reconstruction (red curve) and synthesis (blue curve) distortions in the dataset.}
\label{fig:MOS}
\end{figure}

\begin{figure*}[ht]
\centering
\includegraphics[width=0.85\textwidth]{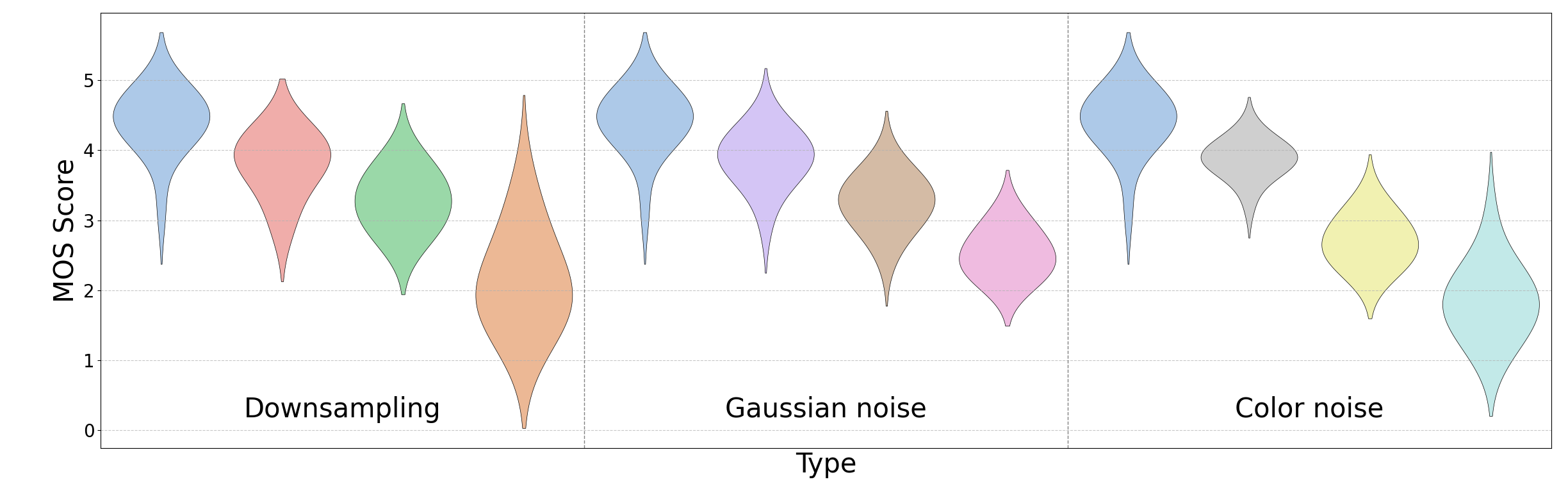}%
\caption{MOS statistics across downsampling, Gaussian noise, and color noise distortions.}
\label{fig:3distort}
\end{figure*}

\subsection{Subjective Data Analysis}

Fig.~\ref{fig:MOS} illustrates the MOS distributions across distortion types. The red curve (reduced viewports and limited training) shows a bimodal pattern, caused by large quality variation—samples with sufficient viewpoints or iterations yield relatively good quality scores, while severely underfitted ones result in poor quality, splitting the score distribution. In contrast, the blue curve (covering downsampling, Gaussian noise, color perturbation) follows a near-normal distribution, as these distortions degrade quality more smoothly and consistently due to uniformly applied parameters. This contrast highlights the differing perceptual impacts of distortion types and supports the reliability of the scoring process.

A more detailed analysis is shown in Fig.~\ref{fig:3distort},  where the relationship between different distortion types and MOS scores is shown for downsampling, Gaussian noise, and color noise. Using the distortion-free model (indicated in blue) as a reference, each distortion type demonstrates a clear and consistent decline in MOS scores as the distortion level increases. This inverse correlation between distortion severity and perceived quality reinforces the theoretical expectation that more severe distortions result in degraded visual quality.

\begin{figure}[ht]
\centering
\includegraphics[width=0.45\textwidth]{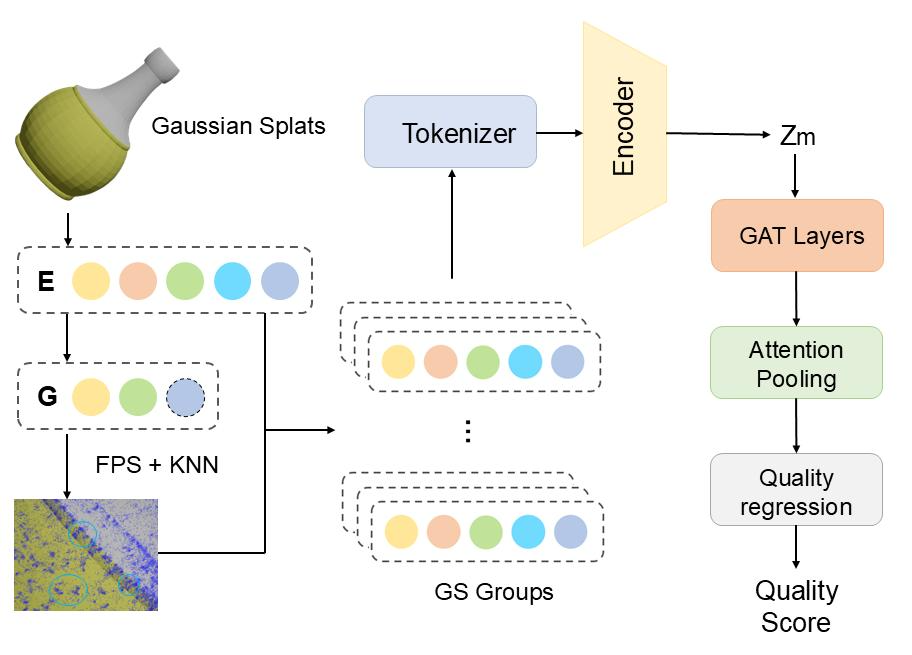}%
\caption{Overall pipeline of the proposed method}
\label{fig:proposed}
\end{figure}

\section{Objective Quality Assessment}
The 3D Gaussian Splatting (3DGS) is represented as a set of $N$ Gaussian primitives: $X = [C, O, S, R, \text{SH}] \in \mathbb{R}^{N \times 59}$, where each Gaussian is encoded with five key attribute groups: centroid coordinates $C \in \mathbb{R}^{N \times 3}$, opacity $O \in \mathbb{R}^{N \times 1}$, spatial scale $S \in \mathbb{R}^{N \times 3}$, rotation represented via quaternion $R \in \mathbb{R}^{N \times 4}$, and appearance descriptors captured using spherical harmonics $\text{SH} \in \mathbb{R}^{N \times 48}$. 

\subsection {The Proposed GSOQA Model}
The proposed 3D Gaussian Splatting Objective Quality Assessment (GSOQA) model aims to assess the perceptual quality of 3DGS in a no-reference manner. The overall framework, illustrated in Fig.~\ref{fig:proposed}, consists of three core modules: geometric preprocessing, semantic feature extraction, and perceptual quality regression.

\subsubsection{Geometric Preprocessing}

To reduce computational overhead, we begin by randomly downsampling the original set $X$ to $p$ Gaussian splats. The attribute set of each primitive is then partitioned into two subspaces: the grouping space $G$ and the embedding space $E$. All attributes are preserved in $E$ to maintain the full descriptive capacity, while $G$ is selectively constructed using $C$, $S$, and the first three coefficients of $\text{SH}$, which together encode key factors—spatial layout, structural scale, and low-frequency appearance—that critically influence visual quality perception.

Using $G$, we apply Farthest Point Sampling (FPS) to select $n$ representative center splats: 
\begin{equation} 
\mathbf{S} = \text{FPS}(G, n). 
\end{equation} 
For each center splat $c_i$, we retrieve its $k$ nearest neighbors within the space $G$, forming a localized perceptual region $\mathbf{I}_{N} \in \mathbb{Z}^{n\times k}$, from which we extract the corresponding high-dimensional embeddings: $E_m = X[\mathcal{S}, \mathbf{I}_{N}, :] \in \mathbb{R}^{n \times k \times 59}$. This preprocessing step enables structured and context-aware perception by enforcing locality and attribute coherence.

\subsubsection{Semantic Feature Extraction}
Although $E_m$ encodes raw descriptive cues, it lacks semantic abstraction necessary for robust perceptual quality inference. To bridge this gap, we adopt a Gaussian-MAE as a pretrained feature extractor. Gaussian-MAE \cite{ma2024shapesplat} is a self-supervised masked autoencoder that learns expressive representations by reconstructing randomly masked splats from latent features. Its reconstruction objective, typically formulated via L1 loss and Chamfer Distance, guides the encoder to capture both fine-grained geometric structures and global appearance priors.

During training, we input the unmasked local embeddings $E_m$ into the encoder of the pretrained Gaussian-MAE to obtain a semantically enhanced feature tensor: 
\begin{equation} 
Z_m = \text{Gaussian-MAE-Encoder}(E_m). 
\end{equation}
This process endows the quality assessment module with high-level perceptual awareness derived from the structure-function learning of Gaussian primitives.

\subsubsection{Perceptual Quality Regression}
The extracted features $H^{(0)} = Z_m$ are refined using a three-stage cascaded Graph Attention Network (GAT) \cite{velivckovic2018graph}, which is designed to model spatial-attribute dependencies within the splat regions. Each GAT block employs a dual-residual architecture that decouples spatial interaction modeling from intra-splat attribute reasoning. The feature propagation is governed by the following update rules:
\begin{align} 
\mathbf{H}^{(l,1)} &= \mathcal{M}\left(\text{LayerNorm}(\mathbf{H}^{(l)})\right) + \mathbf{H}^{(l)}, \\
\mathbf{H}^{(l+1)} &= \mathcal{F}\left(\text{LayerNorm}(\mathbf{H}^{(l,1)})\right) + \mathbf{H}^{(l,1)}, 
\end{align}
where $\mathcal{M}$ and $\mathcal{F}$ denote the GAT and Feedforward Network (FFN) layers, respectively. In the first stage, inter-splat relationships are captured via multi-head graph attention in the spatial domain, enabling the network to learn context-aware embeddings. In the second stage, a point-wise feedforward network with GELU activation captures complex attribute-level interactions within each splat.

To regress a scalar quality score from the refined splat features $H^{(3)}$, we introduce an attribute-aware pooling mechanism. This module dynamically evaluates the contribution of each splat to the overall perceived quality by assigning attention weights: 

\begin{equation}
\mathbf{g}_{\text{feat}} = \sum_{i \in \mathbf{N}_c} \alpha_i \mathbf{h}_i, \quad \alpha_i = \frac{\exp\left( \mathbf{W}_q^T \phi(\mathbf{W}_e^T \mathbf{h}_i) \right)}{\sum_{j \in \mathbf{N}_c} \exp\left( \mathbf{W}_q^T \phi(\mathbf{W}_e^T \mathbf{h}_j) \right)},
\end{equation} 
where $h_i \in H^{(3)}$ is the $d$-dimensional feature vector of the $i$-th splat, $\phi(\cdot)$ denotes GELU activation, $\mathbf{W}_e \in \mathbb{R}^{d \times 2d}$ expands the feature dimension, and $\mathbf{W}_q \in \mathbb{R}^{2d}$ generates the attention scores.

This attention mechanism prioritizes perceptually salient splats, i.e., exhibiting distortions or irregular rendering artifacts, while suppressing the influence of redundant or smooth regions. The aggregated feature $\mathbf{g}_{\text{feat}}$ is then fed into a regression head (implemented as a fully connected layer) to yield the final quality prediction $\hat{y}$.

\subsubsection{Loss Function and Optimization}
To ensure both accuracy and rank consistency, we formulate a hybrid objective function using two complementary loss terms: linearity-induced loss $\mathcal{L}_{lin} = 1 - \frac{\text{Cov}(\hat{y}, y)}{\sigma_{\hat{y}} \sigma_{y}}$, and monotonicity-induced loss $\mathcal{L}_{\text{mon}} = \frac{1}{B^2} \sum_{i=1}^{B} \sum_{j=1}^{B} \mathbb{I}(y_i > y_j) \cdot \max\left(0, 1 - (\hat{y}_i - \hat{y}_j)\right)$, where $B$ is the batch size, $\mathbb{I}(\cdot)$ is the indicator function, $y$ are the true MOS scores, and $\hat{y}$ are the predicted scores. The first term penalizes deviations from linear correlation between predictions and ground-truth MOS, and the second term enforces order-preserving constraints over predicted scores, both of which are calculated over a batch of $M$ samples. The final training loss is defined as: 
\begin{equation}
\min \mathcal{L}_{total} = \lambda_1 \mathcal{L}_{lin} + \lambda_2 \mathcal{L}_{mon},
\end{equation}
with $\lambda_1$ and $\lambda_2$ controlling the trade-off between correlation fidelity and order preservation. These hyperparameters are further discussed in the experimental section.

\subsection{Benchmark Evaluation on 3DGS-QA}\label{sec:experiments}
We evaluate our GSOQA model on the 3DGS-QA dataset to demonstrate its effectiveness in quality assessment.

\subsubsection{Experimental Settings}
To ensure rigor and reproducibility, we employ a standard five-fold cross-validation strategy on the 3DGS-QA dataset. Specifically, the 15 base 3DGS models are randomly divided into five disjoint subsets, each containing three base models. In each fold, the 45 distorted variants associated with one subset are used for testing, while the remaining 180 are used for training. This setup ensures robustness and minimizes bias from any specific split. The model is trained using the AdamW optimizer with a learning rate and weight decay of 1e-4. A OneCycleLR scheduler is applied for learning rate scheduling. Training spans 100 epochs with a batch size of 32. The regularization weights $\lambda_1$ and $\lambda_2$ are both set to 0.5.

\subsubsection{Baselines}
As GSOQA is the first quality assessment model designed specifically for 3D Gaussian primitives, we compare it against ten representative baselines across three categories. 1) Image Quality Assessment (IQA): SSIM~\cite{wang2004image},  BRISQUE~\cite{mittal2012making}, LPIPS~\cite{zhang2018unreasonable},
HyperIQA~\cite{su2020blindly}; 2) Video Quality Assessment (VQA): RAPIQUE~\cite{tu2021rapique}, Speed~\cite{bampis2017speed}, HDRVQM~\cite{narwaria2015hdr}, and FASTVQA~\cite{Wu2022fastvqa}; 3) 3D Quality Assessment (3DQA): PQA-Net~\cite{Liu2021pqanet} and CLIP-PCQA~\cite{liu2025clippcqa}.

\subsubsection{Evaluation Criteria}
We adopt four widely-used statistical metrics to assess model performance: 1) PLCC (Pearson Linear Correlation Coefficient), indicating the strength of linear correlation between predicted scores and subjective ratings. 2) SRCC (Spearman Rank Order Correlation Coefficient), measuring the monotonic relationship between predictions and ground-truth rankings. 3) KRCC (Kendall Rank Correlation Coefficient), assessing ordinal consistency based on concordant and discordant pairs. 4) RMSE (Root Mean Squared Error), quantifying the absolute error between predicted and subjective scores.

\begin{table}[!h]
    \centering
    \caption{Performance comparison on 3DGS-QA database.}
    \renewcommand{\arraystretch}{1.5}
    \label{tab:iqa_classification}
    \resizebox{0.45\textwidth}{!}{
    \begin{tabular}{c|c|c|c|c|c}
        \hline
        Type & Method & PLCC & SRCC & KRCC & RMSE \\
        \hline
        \multirow{4}{*}{\shortstack{IQA}}
        & SSIM & 0.2936 & 0.3117 & 0.2523 & 1.2459 \\
        & BRISQUE & 0.3632 & 0.3330 & 0.2503 & 1.5281 \\
        & LPIPS & 0.3437 & 0.3408 & 0.2740 & 1.6182 \\
        
        & HyperIQA & 0.4253 & 0.3059 & 0.2335 & 1.3141 \\
        \hline
        \multirow{3}{*}{\shortstack{VQA} } 
        & RAPIQUE & 0.3803 & 0.3077 & 0.2045 & 1.7214 \\
        & Speed & 0.2410 & 0.2805 & 0.2073 & 2.1454 \\
        & HDRVQM & 0.4419 & 0.3354 & 0.2351 & 2.6294 \\
        & FASTVQA & 0.3320 & 0.2886 & 0.2234 & 1.7284 \\
        \hline
        \multirow{3}{*}{\shortstack{3DQA} } 
        & PQA-Net & 0.4896 & 0.4168 & 0.3873 & 1.1457 \\
        & CLIP-PCQA & 0.5698 & 0.6215 & 0.4478 & 1.2659 \\
        & Proposed & \textbf{0.8213} & \textbf{0.7738} & \textbf{0.5896} & \textbf{1.0798} \\
        \hline
    \end{tabular}
    }
\end{table}

\begin{table*}[!ht]
    \centering
    \caption{Performance comparison of different methods on 3DGS-QA database across distortion types.}
    \renewcommand{\arraystretch}{1.5}
    \label{tab:iqa_classification_distortion}
    \resizebox{0.95\textwidth}{!}{
    \begin{tabular}{c|ccc|ccc|ccc|ccc}
        \hline
        \multirow{2}{*}{Method} & \multicolumn{3}{c|}{Reconstruction Distortion} & \multicolumn{3}{c|}{Downsampling} & \multicolumn{3}{c|}{Gaussian Noise} & \multicolumn{3}{c}{Color Noise} \\
        \cline{2-4} \cline{5-7} \cline{8-10} \cline{11-13}
        & PLCC & SRCC & KRCC & PLCC & SRCC & KRCC & PLCC & SRCC & KRCC & PLCC & SRCC & KRCC \\
        \hline
        SSIM & 0.3723 & 0.4609 & 0.3537 & 0.5069 & 0.5202 & 0.3647 & 0.5506 & 0.5324 & 0.3725 & 0.5241 & 0.4858 & 0.3344 \\
        BRISQUE & 0.3145 & 0.3133 & 0.2538 & 0.6667 & 0.5664 & 0.3334 & 0.4345 & 0.4904 & 0.3766 & 0.4574 & 0.5314 & 0.3867 \\
        LPIPS & 0.3827 & 0.4172 & 0.2813 & 0.4150 & 0.4724 & 0.3162 & 0.5302 & 0.5194 & 0.3632 & 0.4981 & 0.4814 & 0.3321 \\
        HyperIQA & 0.5140 & 0.4854 & 0.3411 & 0.3950 & 0.2028 & 0.1258 & 0.3624 & 0.3750 & 0.2814 & 0.4759 & 0.5371 & 0.4267 \\
        RAPIQUE & 0.4055 & 0.3652 & 0.2405 & 0.4894 & 0.4285 & 0.2803 & 0.4350 & 0.3855 & 0.2608 & 0.4605 & 0.3956 & 0.2504 \\
        Speed & 0.2651 & 0.2955 & 0.2202 & 0.3305 & 0.3550 & 0.2501 & 0.2857 & 0.3105 & 0.2305 & 0.3005 & 0.3252 & 0.2403 \\
        HDRVQM  & 0.4605 & 0.3908 & 0.2612 & 0.5312 & 0.4505 & 0.3005 & 0.4805 & 0.4126 & 0.2809 & 0.5102 & 0.4305 & 0.2906 \\
        FASTVQA & 0.3517 & 0.3246 & 0.2287 & 0.4026 & 0.3714 & 0.2527 & 0.3816 & 0.3329 & 0.2416 & 0.3921 & 0.3485 & 0.2654 \\
        PQA-Net & 0.5282 & 0.4926 & 0.3815 & 0.5617 & 0.4983 & 0.3654 & 0.5126 & 0.4752 & 0.3726 & 0.5354 & 0.4821 & 0.4865 \\
        CLIP-PCQA & 0.6151 & 0.6725 & 0.4632 & 0.6485 & 0.6933 & 0.4525 & 0.5863 & 0.6341 & 0.4556 & 0.6239 & \textbf{0.5884} & \textbf{0.4985} \\
        Proposed & \textbf{0.8694} & \textbf{0.6732} & \textbf{0.5374} & \textbf{0.8717} & \textbf{0.7622} & \textbf{0.6061} & \textbf{0.7164} & \textbf{0.6723} & \textbf{0.4573} & \textbf{0.7083} & 0.3833 & 0.2222 \\
        \hline
    \end{tabular}
    }
\end{table*}    

\subsubsection{Experimental Results and Discussion}
We conduct extensive experiments on the 3DGS-QA database to evaluate the effectiveness of GSOQA. As shown in Table~\ref{tab:iqa_classification}, our model significantly outperforms all compared IQA, VQA, and 3DQA approaches across all evaluation metrics.  To ensure fair comparisons with subjective perception, IQA/VQA models were applied to rendered video sequences from 40 uniformly distributed viewpoints. For IQA models, predictions are averaged across individual frames, while VQA models assess the sequences as a whole. 

The results reveal clear limitations of existing quality assessment models in handling the unique characteristics of 3DGS content. Specifically, conventional IQA metrics fail to capture the complex perceptual artifacts introduced by the synthesis process, as evidenced by their low PLCC values (all below 0.45) and high RMSE scores (exceeding 1.2). VQA methods similarly underperform, largely due to their inability to account for depth-related distortions and structural ambiguities inherent in 3D-generated views. Traditional 3DQA methods also struggle, as they rely on explicit geometric representations (e.g., point clouds), which are fundamentally incompatible with the probabilistic and continuous nature of 3D Gaussian Splatting. In contrast, the proposed GSOQA model directly operates on native Gaussian primitives, enabling it to better capture both structural integrity and perceptual quality of 3DGS renderings. Its superior performance across all evaluation metrics underscores the benefit of learning quality representations from the original, physically meaningful data.

Further insights are provided by the individual distortion analysis in Table~\ref{tab:iqa_classification_distortion}. GSOQA always achieves the highest PLCC and SRCC scores across all four distortion categories: reconstruction artifacts, downsampling, Gaussian noise, and color noise. Notably, in the cases of reconstruction and downsampling—two distortion types that frequently occur in real-world rendering pipelines—GSOQA obtains PLCC scores above 0.85 and SRCC values exceeding 0.65, significantly outperforming all baselines. These results highlight the model’s robustness and its ability to accurately reflect perceptual quality degradation caused by both fidelity loss and data noise in 3DGS.

\begin{table}[ht]
\caption{Ablation analysis on the 3DGS-QA database.}
    \label{tab:Ablation}
    \renewcommand{\arraystretch}{1.5}
    \resizebox{0.45\textwidth}{!}{
\begin{tabular}{c|c|c|c|c}
        \hline
Model & PLCC   & SRCC   & KRCC   & RMSE   \\
\hline
w/o GAT & 0.7914 & 0.7496 & 0.5591 & 1.2285 \\
w/o attention pooling & 0.7952 & 0.7359 & 0.5551 & 1.2651 \\
w/o $\mathcal{L}_{mon}$ & 0.8046 & 0.7124 & 0.5003 & 1.2867 \\
w/o $\mathcal{L}_{lin}$ & 0.7161 & 0.7954 & 0.5043 & 1.5948 \\
Full model & 0.8213 & 0.7738 & 0.5896 & 1.0798 \\
        \hline
    \end{tabular}
    }
\end{table}

\subsubsection{Ablation Study}
To evaluate the individual contributions of key components within the GSOQA framework, we perform a series of ablation experiments, with results summarized in Table~\ref{tab:Ablation}.

Removing the GAT module results in a noticeable decline in PLCC (from 0.8213 to 0.7914) and SRCC (from 0.7738 to 0.7496), highlighting the importance of adaptively modeling relational dependencies among Gaussian primitives for accurate perceptual quality prediction. Additionally, replacing the perceptual attention pooling with a naive global pooling strategy leads to performance degradation across all evaluation metrics, particularly in SRCC (dropping to 0.7359). This validates the effectiveness of the attention mechanism in emphasizing perceptually salient regions, such as distorted or irregular primitives.

Moreover, the removal of regularization terms that enforce monotonicity or linearity also adversely impacts model performance, revealing their complementary roles in promoting perceptual consistency. Notably, the absence of $\mathcal{L}_{mon}$ (i.e., $\lambda_2 = 0$) significantly reduces SRCC, while removing $\mathcal{L}_{lin}$ (i.e., $\lambda_1 = 0$) causes the largest reduces PLCC, suggesting its importance for stabilizing regression outputs. Additional experiments under varying combinations of $\lambda_1$ and $\lambda_2$, reported in Table~\ref{tab:combine}, further validate the complementary nature of these regularization terms in promoting robust and perceptually aligned quality prediction.

\begin{table}[ht]
\tiny
\caption{The analysis on the parameters $\lambda_1$ and $\lambda_2$.}
    \label{tab:combine}
    \renewcommand{\arraystretch}{1.5}
    \resizebox{0.45\textwidth}{!}{
\begin{tabular}{c|c|c|c|c|c}
        \hline
$\lambda_1$ & $\lambda_2$ & PLCC & SRCC   & KRCC   & RMSE   \\
\hline
0.2 & 0.8 & 0.8002 & 0.7828 & 0.5177 & 1.1311 \\
0.4 & 0.6 & 0.8244 & 0.7653 & 0.5383 & 1.0647 \\
0.6 & 0.4 & 0.7882 & 0.7415 & 0.5412 & 1.1185\\
0.8 & 0.2 & 0.7953 & 0.7408 & 0.5405 & 1.1302\\
        \hline
    \end{tabular}
    }
\end{table}


\section{Conclusion}
In this work, we conducted the first dedicated study on perceptual quality assessment for 3D Gaussian Splatting (3DGS). To this end, we constructed 3DGS-QA, a comprehensive subjective quality dataset that includes diverse object categories and degradation types, enabling systematic analysis of perceptual artifacts in 3DGS-rendered content. Furthermore, we proposed a novel no-reference assessment model that directly analyzes native 3D Gaussian primitives to predict perceptual quality without relying on 2D projections or ground-truth references. Extensive experiments demonstrate the superior performance of our approach compared to existing multimedia quality techniques. Future efforts will explore generalization to complex dynamic scenes and integration with rendering optimization pipelines. 


\section{Acknowledgment} 
This work was supported by the National Key R\&D Program of China (2023YFA1008500), the National Natural Science Foundation of China (NSFC) under grant U22B2035, the Natural Science Foundation of Heilongjiang under Grant LH2024F018, the China Post-Doctoral Science Foundation under Grant 2023M740939, the Heilongjiang Post-Doctoral Foundation under Grant LBHZ23170, and the Key Laboratory of Cognitive Intelligence and Content Security, Ministry of Education (Grant no. RZZN202501, Harbin Institute of Technology).
\bibliography{main}

\end{document}